\documentclass[final]{cvpr}
\usepackage{times}
\usepackage{epsfig}
\usepackage{graphicx}
\usepackage{amsmath}
\usepackage{amssymb}
\usepackage{subfigure}
\usepackage{multirow}
\usepackage[hang,flushmargin]{footmisc}
\usepackage{pifont}
\usepackage{xcolor}

\def\cvprPaperID{***} 
\def\confYear{CVPR 2021}

\usepackage[pagebackref=true,breaklinks=true,letterpaper=true,colorlinks,bookmarks=false]{hyperref}
\newcommand{\rf}[1]{{\textbf{\color{red}{#1}}}} 
\newcommand{\bd}[1]{{\underline{\color{blue}{#1}}}} 
\newcommand{\cmark}{\ding{51}}%
\newcommand{\xmark}{\ding{55}}%
\definecolor{amethyst}{rgb}{0.6, 0.4, 0.8}

\begin{document}

\title{BasicVSR: The Search for Essential Components in \\ Video Super-Resolution and Beyond}

\author{Kelvin C.K. Chan$^{1}$
\quad
Xintao Wang$^{2}$
\quad
Ke Yu$^{3}$
\quad
Chao Dong$^{4,5}$
\quad
Chen Change Loy$^{1}$\thanks{Corresponding author}\\
$^{1}$S-Lab, Nanyang Technological University\quad
$^{2}$Applied Research Center, Tencent PCG\\
$^{3}$CUHK – SenseTime Joint Lab, The Chinese University of Hong Kong\\
$^{4}$Shenzhen Key Lab of Computer Vision and Pattern Recognition, SIAT-SenseTime Joint Lab,\\Shenzhen Institutes of Advanced Technology, Chinese Academy of Sciences\\
$^{5}$SIAT Branch, Shenzhen Institute of Artificial Intelligence and Robotics for Society\\
{\tt\small \{chan0899, ccloy\}@ntu.edu.sg \hspace{1cm} xintao.wang@outlook.com}\\
{\tt\small yk017@ie.cuhk.edu.hk\hspace{1cm}chao.dong@siat.ac.cn}\\\vspace{-1cm}
}
\maketitle
\pagestyle{empty}
\thispagestyle{empty}

\begin{abstract}
	Video super-resolution (VSR) approaches tend to have more components than the image counterparts as they need to exploit the additional temporal dimension. Complex designs are not uncommon. In this study, we wish to untangle the knots and reconsider some most essential components for VSR guided by four basic functionalities, i.e., Propagation, Alignment, Aggregation, and Upsampling. By reusing some existing components added with minimal redesigns, we show a succinct pipeline, BasicVSR, that achieves appealing  improvements in terms of speed and restoration quality in comparison to many state-of-the-art algorithms. We conduct systematic analysis to explain how such gain can be obtained and discuss the pitfalls. We further show the extensibility of BasicVSR by presenting an information-refill mechanism and a coupled propagation scheme to facilitate information aggregation. The BasicVSR and its extension, IconVSR, can serve as strong baselines for future VSR approaches.
\end{abstract}
\vspace{-0.45cm}


%

\begin{table*}[!t]
	\caption{\textbf{Components in existing VSR methods}. We categorize components based on their functionalities: i) \textit{Propagation} refers to the way in which features are propagated temporally, ii) \textit{Alignment} concerns on the spatial transformation applied to misaligned images/features, iii)  \textit{Aggregation} defines the steps to combine aligned features, and iv) \textit{Upsampling} describes the method to transform the aggregated features to the final output image. Bolded texts correspond to designs that were reported to achieve better performance in the literature.}
	\label{tab:summary}
	\begin{center}
		\tabcolsep=0.1cm
		\scalebox{0.77}{
			\begin{tabular}{l|c|c|c|c|c|c|c|c}
				\hline
				\multirow{2}{*}{} & \multicolumn{3}{c|}{Sliding-Window} & \multicolumn{5}{c}{Recurrent}                                                                                                                                                                                                            \\ \cline{2-9}
				                  & EDVR~\cite{wang2019deformable}      & MuCAN~\cite{li2020mucan}      & TDAN~\cite{tian2018tdan} & BRCN~\cite{huang2015bidirectional,huang2018video} & FRVSR~\cite{sajjadi2018frame} & RSDN~\cite{isobe2020video1} & \textbf{\mbox{BasicVSR}} & \textbf{\mbox{IconVSR}}          \\ \hline
				Propagation       & Local                               & Local                         & Local                    & \textbf{Bidirectional}                            & Unidirectional                & Unidirectional              & \textbf{Bidirectional}   & \textbf{Bidirectional} (coupled) \\
				Alignment         & \textbf{Yes} (DCN)                  & \textbf{Yes} (correlation)    & \textbf{Yes} (DCN)       & No                                                & \textbf{Yes} (flow)           & No                          & \textbf{Yes} (flow)      & \textbf{Yes} (flow)              \\
				Aggregation       & Concatenate + \textbf{TSA}          & Concatenate                   & Concatenate              & Concatenate                                       & Concatenate                   & Concatenate                 & Concatenate              & Concatenate + \textbf{Refill}    \\
				Upsampling        & Pixel-Shuffle                       & Pixel-Shuffle                 & Pixel-Shuffle            & Pixel-Shuffle                                     & Pixel-Shuffle                 & Pixel-Shuffle               & Pixel-Shuffle            & Pixel-Shuffle                    \\ \hline
			\end{tabular}}
		\vspace{-0.3cm}
	\end{center}
\end{table*}

\section{Introduction}
Compared to single-image super-resolution, which focuses on the intrinsic properties of a single image for the upscaling task, video super-resolution (VSR) poses an extra challenge as it involves aggregating information from multiple highly-related but misaligned frames in video sequences.

\begin{figure}[t]
	\begin{center}
		\includegraphics[width=0.49\textwidth]{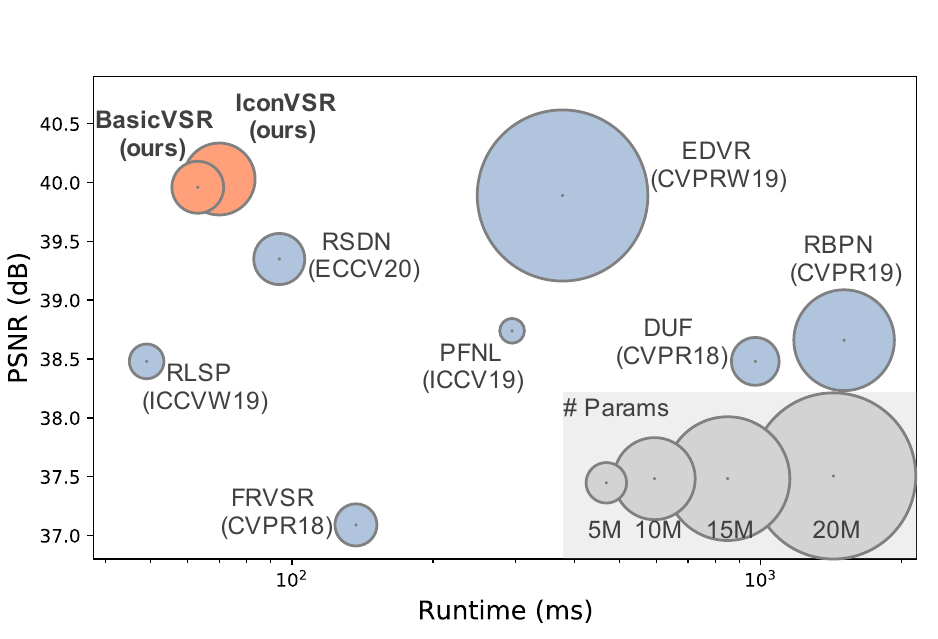}
		\caption{\textbf{Speed and performance comparison.} Without bells and whistles, \mbox{BasicVSR} outperforms state-of-the-art methods with high efficiency. Built upon \mbox{BasicVSR}, \mbox{IconVSR} further improves the performance. Comparisons are performed on UDM10 dataset~\cite{yi2019progressive}.}
		\vskip -0.7cm
		\label{fig:teaser}
	\end{center}
\end{figure}

Various approaches have been proposed to address the challenge. Some designs can be highly complex. For instance, in the representative method EDVR~\cite{wang2019edvr}, a multi-scale deformable alignment module and multiple attention layers are adopted for aligning and integrating the features from different frames. In RBPN~\cite{haris2019recurrent}, multiple projection modules are used to sequentially aggregate features from multiple frames.
Such designs are effective but inevitably increase the runtime and model complexity (see Fig.~\ref{fig:teaser}).
In addition, unlike SISR, the potentially complex and dissimilar designs of VSR methods pose difficulties in implementing and extending existing approaches, hampering reproducibility and fair comparisons.

There is a need to step back and reconsider the diverse designs of VSR models, with the aim to search for a more generic, efficient, and easy-to-implement baseline for VSR.
We start our search by decomposing popular VSR approaches into submodules based on functionalities. As summarized in Table~\ref{tab:summary}, most existing methods entail four inter-related components, namely, \textit{propagation, alignment, aggregation, and upsampling}.
Such a decomposition allows us to systematically study various options under each component and understand their pros and cons.

Through extensive experiments, we find that with minimal redesigns of existing options, one could already reach a strong yet efficient baseline for VSR without bells and whistles.
In this paper, we highlight one of such possibilities, named \textbf{\mbox{BasicVSR}}.
We observe that, among the four aforementioned components, the choices of propagation and alignment components could lead to a big swing in terms of performance and efficiency.
Our experiments suggest the use of bidirectional propagation scheme to maximize information gathering, and an optical flow-based method to estimate the correspondence between two neighboring frames for feature alignment.
By simply streamlining these propagation and alignment components with the commonly-adopted designs for aggregation (\ie feature concatenation) and upsampling (\ie pixel-shuffle~\cite{shi2016real}), \mbox{BasicVSR} outperforms existing state of the arts~\cite{haris2019recurrent,isobe2020video1,wang2019edvr} in both performance (up to 0.61~dB) and efficiency (up to $24{\times}$ speedup).

Thanks to its simplicity and versatility, \mbox{BasicVSR} provides a viable starting point for extending to more elaborated networks.
By using \mbox{BasicVSR} as a foundation, we present \textbf{\mbox{IconVSR}} that comprises two novel extensions to improve the aggregation and the propagation components.
The first extension is named \textit{information-refill}. This mechanism leverages an additional module to extract features from sparsely selected frames (keyframes), and the features are then inserted into the main network for feature refinement.
The second extension is a \textit{coupled propagation} scheme, which facilitates information exchange between the forward and backward propagation branches.
The two modules not only reduce error accumulation during propagation due to occlusions and image boundaries, but also allow the propagation to access complete information in a sequence for generating high-quality features.
With these two new designs, \mbox{IconVSR} surpasses \mbox{BasicVSR} with a PSNR improvement of up to 0.31 dB.

We believe that our work is timely, given the increasing number of approaches centered around the research of VSR. A strong, simple yet extensible baseline is needed.
Guided by the main functionalities in VSR approaches, we reconsider some essential components in existing pipelines and present an efficient baseline for VSR. We show that simple components, when integrated properly, would synergize and lead to state-of-the-art performance.
We further present an example of extending \mbox{BasicVSR} with two novel modules to refine the propagation and aggregation components.

%
\begin{figure*}[!t]
	\begin{center}
		\includegraphics[width=0.99\textwidth]{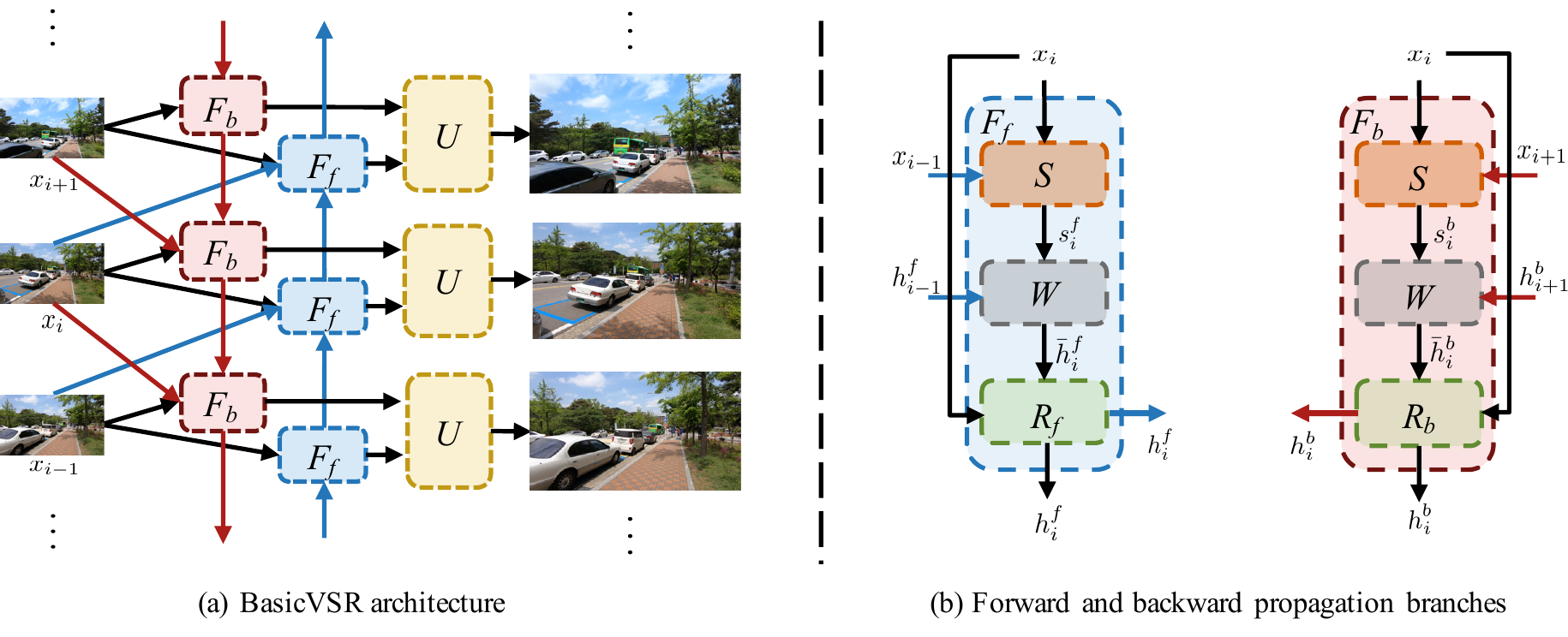}\vspace{0.1cm}
		\caption{\textbf{An overview of \mbox{BasicVSR}.} \mbox{BasicVSR} is a generic and efficient baseline for VSR. With minimal redesigns of existing components including optical flow and residual blocks, it outperforms existing state of the arts with high efficiency. \textbf{(a)} \mbox{BasicVSR} adopts a typical bidirectional recurrent network. The upsampling module $U$ contains multiple pixel-shuffle and convolutions. The red and blue colors represent the backward and forward propagations, respectively. \textbf{(b)} The propagation branches contain only generic components. $S$, $W$, and $R$ refer to the flow estimation module, spatial warping module, and residual blocks, respectively.}
		\vspace{-0.5cm}
		\label{fig:overview}
	\end{center}
\end{figure*}

\section{Related Work}
\label{sec:relatedwork}
Existing VSR approaches~\cite{huang2015bidirectional,liu2014bayesian,takeda2009super,yi2019progressive,li2020mucan,isobe2020video1,isobe2020video} can be mainly divided into two frameworks -- \textit{sliding-window} and \textit{recurrent}. Earlier methods~\cite{caballero2017real,tao2017detail,xue2019video} in the sliding-window framework predict the optical flow between low-resolution (LR) frames and perform spatial warping for alignment. Later approaches resort to a more sophisticated approach of implicit alignment. For example, TDAN~\cite{tian2018tdan} adopts deformable convolutions (DCNs)~\cite{dai2017deformable,zhu2019deformable} to align different frames at the feature level. EDVR~\cite{wang2019edvr} further uses DCNs in a multi-scale fashion for more accurate alignment. DUF~\cite{jo2018deep} leverages dynamic upsampling filters to handle motions implicitly.
Some approaches take a recurrent framework. RSDN~\cite{isobe2020video1} proposes a recurrent detail-structural block and a hidden state adaptation module to enhance the robustness to appearance change and error accumulation. RRN~\cite{isobe2020revisiting} adopts a residual mapping between layers with identity skip connections to ensure a fluent information flow and preserve the texture information over long periods.
The aforementioned studies have led to many new and sophisticated components to address the propagation and alignment problems in VSR. Here, we reinvestigate some of the components and find that bidirectional propagation coupled with a simple optical flow-based feature alignment suffice to outperform many state-of-the-art methods.

The information-refill mechanism in \mbox{IconVSR} is reminiscent of the concept of interval-based processing~\cite{chen2018optimizing,jain2019accel,shelhamer2016clockwork,zhang2017fast,zhu2018towards,zhu2017flow,zhu2017deep}. These methods divide video frames into independent intervals characterized by keyframes and non-keyframes. The keyframes and non-keyframes are then processed by different pipelines. For instance, FAST~\cite{zhang2017fast} applies SRCNN~\cite{dong2014learning,dong2016image} to super-resolve the keyframes. Non-keyframes are then restored using the upscaled keyframes and the motion vectors stored in the compressed video codec.
\mbox{IconVSR} inherits the concept of keyframes, but unlike existing methods that process the intervals independently, we make one advancement by connecting the intervals through the propagation branches. With this design, long-term information can be propagated across the inter-connected intervals, further improving the effectiveness.

%
\section{Methodology}
Video super-resolution, by nature, involves a long and complex processing pipeline since it needs to aggregate information from not only the spatial dimension but also the temporal dimension. Existing studies typically focus on one aspect of the functionalities to make advancement and may not collectively consider the synergy of various components.
There is an urge to revisit various components macroscopically and uncover a generic baseline that inherits the strengths of existing approaches. In this work, we conduct extensive analysis and present a simple, strong and versatile baseline, BasicVSR, which can serve as a backbone with abundant flexibilities in design.

\subsection{\mbox{BasicVSR}}
\label{sec:basicvsr}
Aiming at discovering generic frameworks for facilitating analysis and development of VSR methods, we confine our search to commonly-adopted elements such as optical flow and residual blocks. An overview of \mbox{BasicVSR} is depicted in Fig.~\ref{fig:overview}.

\begin{figure}[!t]
	\begin{center}
		\includegraphics[width=0.45\textwidth]{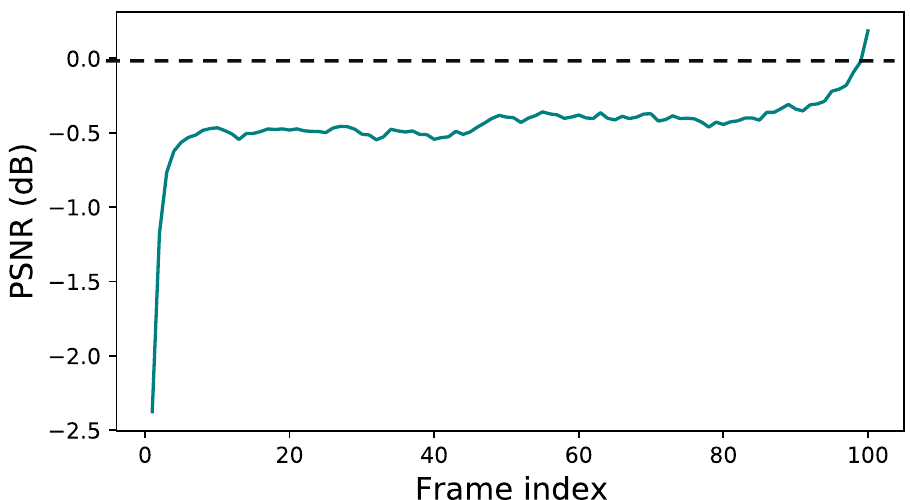}
		\caption{\textbf{Unidirectional vs Bidirectional.} In unidirectional propagation, earlier timesteps receive less information, leading to inferior performance. Values smaller than zero (dotted line) indicates a lower PSNR than the bidirectional counterpart. Note that the unidirectional model outperforms the bidirectional model only for the last frame, owing to the zero feature initialization in the bidirectional model.}
		\label{fig:unidirectional}
	\end{center}
	\vspace{-0.5cm}
\end{figure}

\vspace{0.2cm}
\noindent\textbf{Propagation.}
Propagation is one of the most influential components in VSR. It specifies how the information in a video sequence is leveraged. Existing propagation schemes can be divided into three main groups: \textit{local}, \textit{unidirectional} and \textit{bidirectional} propagations. In what follows, we discuss the weaknesses of the former two to motivate our choice of bidirectional propagation in BasicVSR.
\begin{itemize}
	\item \textbf{\textit{Local Propagation.}}
	      The sliding-window methods~\cite{haris2019recurrent,isobe2020video,wang2019edvr} take the LR images within a local window as inputs and employ the local information for restoration. In this design, the accessible information is restricted in a local neighborhood. The omittance of distant frames inevitably limits the potential of the sliding-window methods. To verify our claim, we start with a global receptive field (in the temporal dimension) and gradually reduce the receptive field. We separate the test sequences into $K$ segments and use our \mbox{BasicVSR} to restore each segment independently. The PSNR difference to the case $K{=}1$ (global propagation) is depicted in Fig.~\ref{fig:segment}.

	      First, the difference in PSNR is reduced (\ie~better performance) when the number of segments decreases (\ie~temporal receptive field increases). This suggests that the information in distant frames is beneficial to the restoration and should not be neglected. Second, the difference in PSNR is the largest at the two ends of each segment, indicating the necessity of adopting long sequences to accumulate long-term information.\vspace{0.3cm}
	      \begin{figure}[!t]
		      \begin{center}
			      \includegraphics[width=0.45\textwidth]{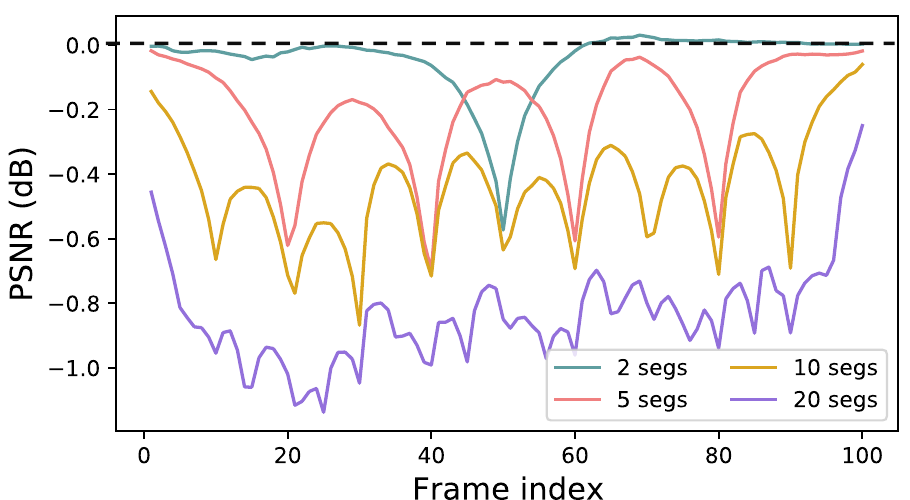}
			      \caption{\textbf{Local vs Global propagation.} When the number of segments $K$ is reduced, the increased temporal receptive field leads to higher PSNR. This demonstrates the importance of aggregating long-term information. Values smaller than zero (dotted line) indicate a lower PSNR than the case $K{=}1$.}
			      \vspace{-0.4cm}
			      \label{fig:segment}
		      \end{center}
	      \end{figure}
	\item \textit{\textbf{Unidirectional Propagation.}}
	      The aforementioned problem can be resolved by adopting a unidirectional propagation~\cite{fuoli2019efficient,isobe2020video1,isobe2020revisiting,sajjadi2018frame}, where the information is sequentially propagated from the first frame to the last frame. However, in this setting, the information received by different frames is imbalanced. Specifically, the first frame receives no information from the video sequence except itself, whereas the last frame receives information from the whole sequence. Hence, suboptimal results are expected for the earlier frames.

	      To demonstrate the effects, we compare BasicVSR (using bidirectional propagation) with its unidirectional variant (with comparable network complexity). From Fig.~\ref{fig:unidirectional}, we see that the unidirectional model obtains a significantly lower PSNR than bidirectional propagation at early timesteps, and the difference gradually reduces as more information is aggregated with the increase in the number of frames. Moreover, a consistent performance drop of 0.5~dB is observed with only partial information employed.
	      These observations reveal the suboptimality of unidirectional propagation. One can improve the output quality by propagating information back from the last frame of the sequence.
	\item \textit{\textbf{Bidirectional Propagation.}}
	      The above two problems can be simultaneously addressed by bidirectional propagation, in which the features are propagated forward and backward in time independently. Motivated by this, \mbox{BasicVSR} adopts a typical bidirectional propagation scheme.
	      Given an LR image $x_i$, its neighboring frames $x_{i-1}$ and $x_{i+1}$, and the corresponding features propagated from its neighbors, denoted as $h^f_{i-1}$ and $h^b_{i+1}$, we have
	      \begin{equation}
		      \label{eq:pipeline_basicvsr}
		      \begin{split}
			      &h_i^b = F_b(x_i, x_{i+1}, h_{i+1}^b),\\
			      &h_i^f = F_f(x_i, x_{i-1}, h_{i-1}^f),
		      \end{split}
	      \end{equation}
	      where $F_b$ and $F_f$ denote the backward and forward propagation branches, respectively.
\end{itemize}

\noindent\textbf{Alignment.} Spatial alignment plays an important role in VSR as it is responsible to align highly related but misaligned images/features for subsequent aggregation. Mainstream works can be divided into three categories: \textit{without alignment}, \textit{image alignment}, and \textit{feature alignment}. In this section, we conduct experiments to analyze each of the categories and to validate our choice of feature alignment.

\begin{itemize}
	\item \textit{\textbf{Without Alignment.}}
	      Existing recurrent methods~\cite{fuoli2019efficient,huang2015bidirectional,huang2018video,isobe2020video1,isobe2020revisiting} generally do not perform alignment during propagation. The non-aligned features/images impede aggregation and eventually lead to substandard performance. This suboptimality can be reflected by our experiment, where we remove the spatial alignment module in BasicVSR. In this case, we directly concatenate the non-aligned features for restoration. Without proper alignment, the propagated features are not spatially aligned with the input image. As a result, the local operations such as convolutions, which have relatively small receptive fields, are inefficient in aggregating the information from corresponding locations. A drop of 1.19~dB of PSNR is observed. This result suggests that it is pivotal to adopt operations that have a large enough receptive field to aggregate information from distant spatial locations.
	\item \textit{\textbf{Image Alignment.}}
	      Earlier works~\cite{kim2018spatio,xue2019video} perform alignment by computing the optical flow and warping the images before restoration. Recently, Chan~\etal~\cite{chan2020understanding} show that moving the spatial alignment from the image level to the feature level yields a marked improvement. In this work, we further conduct experiments to verify their claim. We compare image warping and feature warping\footnote{We compute optical flow from the images and use the optical flow for feature warping.} on a variant of BasicVSR. Resulting from the inaccuracy of optical flow estimation, the warped images inevitably suffer from blurriness and incorrectness. The loss of details eventually leads to degraded outputs. In our experiments, a drop of 0.17~dB is observed when adopting image alignment. This observation confirms the necessity of shifting the spatial alignment to the feature level.
	\item \textit{\textbf{Feature Alignment.}}
	      The inferior performance of removing/image alignment motivates us to resort to feature alignment. Similar to flow-based methods~\cite{kim2018spatio,sajjadi2018frame,xue2019video}, BasicVSR adopts optical flow for spatial alignment. But instead of warping the images as in previous works, we perform warping on the features for better performance. The aligned features are then passed to multiple residual blocks for refinement.
	      Formally, we have
	      \begin{equation}
		      \label{eq:prop_branch}
		      \begin{split}
			      &s_i^{\{b,f\}} = S(x_i, x_{i\pm1}),\\
			      &\bar{h}_i^{\{b,f\}} = W(h_{i\pm1}^{\{b,f\}}, s_i^{\{b,f\}}),\\
			      &h_i^{\{b,f\}} = R_{\{b,f\}}(x_i, \bar{h}_i^{\{b,f\}}),
		      \end{split}
	      \end{equation}
	      and $F_{\{b,f\}} = R_{\{b,f\}} \circ W \circ S$ with a slight abuse of notations. Here $S$ and $W$ denote the flow estimation and spatial warping modules, respectively, and $R_{\{b,f\}}$ denotes a stack of residual blocks.
\end{itemize}

\noindent\textbf{Aggregation and Upsampling.}
BasicVSR adopts basic components for aggregation and upsampling.
Specifically, given the intermediate features $h_i^{\{b,f\}}$, an upsampling module composed of multiple convolutions and pixel-shuffle~\cite{shi2016real} is used to generate the output HR images:
\begin{equation}
	\label{eq:upsample}
	\begin{split}
		&y_i = U(h_i^f, h_i^b),
	\end{split}
\end{equation}
where $U$ denotes the upsampling module.

\vspace{0.15cm}
\noindent\textbf{Summary of \mbox{BasicVSR}.}
The analysis above motivates the design choice of \mbox{BasicVSR}. For propagation, BasicVSR has chosen bidirectional propagation with emphasis on long-term and global propagation. For alignment, BasicVSR adopts a simple flow-based alignment but  taking place at feature level. For aggregation and upsampling, popular choices on feature concatenation and pixel-shuffle suffice.
Despite being a simple and succinct method, \mbox{BasicVSR} achieves great performance in both restoration quality and efficiency. \mbox{BasicVSR} is also highly versatile as it can readily accommodate additional components to handle more challenging scenarios, as we show next.
\begin{figure}[!t]
	\begin{center}
		\includegraphics[width=0.49\textwidth]{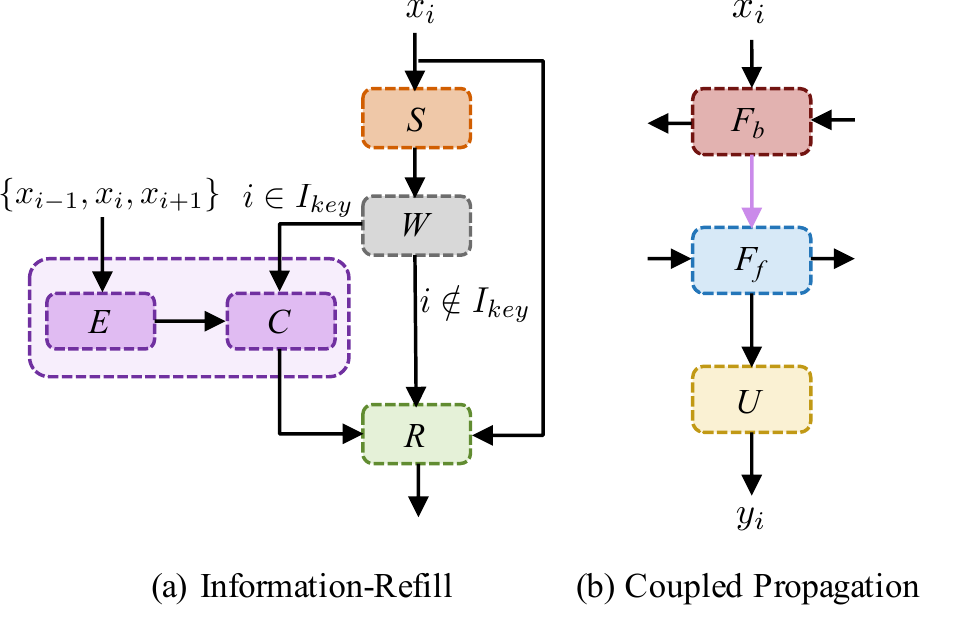}
		\caption{\textbf{(a)} An additional feature extractor is used for feature refinement, alleviating the error accumulation during propagation. $I_{key}$ denotes the set of indices of the selected keyframes. $E$ and $C$ denote the feature extractor and convolution, respectively. \textbf{(b)} The inter-connected propagation branches facilitate the information exchange by passing the outputs of the backward branches to the forward branches. The proposed components are colored in \color{amethyst}{purple}.}
		\vspace{-0.5cm}
		\label{fig:iconvsr}
	\end{center}
\end{figure}
\subsection{From \mbox{BasicVSR} to \mbox{IconVSR}}
Using \mbox{BasicVSR} as a backbone, we introduce two novel components -- \textit{\textbf{I}nformation-refill mechanism} and \textit{\textbf{co}upled propagatio\textbf{n}} (\mbox{\textbf{IconVSR}}), to mitigate error accumulation during propagation and to facilitate information aggregation.
\vspace{0.15cm}

\noindent\textbf{Information-Refill.}
Inaccurate alignment in occluded regions and on image boundaries is a prominent challenge that can lead to error accumulation, especially if we adopt long-term propagation in our framework. To alleviate undesirable effects brought by such erroneous features, we propose an information-refill mechanism for feature refinement.

As shown in Fig.~\ref{fig:iconvsr}(a), an additional feature extractor is used to extract deep features from a subset of input frames (keyframes) and their respective neighbors. The extracted features are then fused with the aligned features $\bar{h}_i$ (Eq.~\ref{eq:prop_branch}) by a convolution:
\begin{equation}
	\label{eq:irblock-detail}
	\begin{split}
		&e_i = E(x_{i-1}, x_i,  x_{i+1}),\\
		&\hat{h}_i^{\{b,f\}} =
		\begin{cases}
			C\left(e_i, \bar{h}_i^{\{b,f\}}\right) & \text{if }i\in I_{key}, \\\\
			\bar{h_i}^{\{b,f\}}                    & \text{otherwise},
		\end{cases}\\
	\end{split}
\end{equation}
where $E$ and $C$ correspond to the feature extractor and convolution, respectively. $I_{key}$ denotes the set of indices of the selected keyframes.
The refined features are then passed to the residual blocks for further refinement:
\begin{equation}
	h_i^{\{b,f\}} = R_{\{b,f\}}(x_i, \hat{h}_i^{\{b,f\}}).
\end{equation}
It is noteworthy that the feature extractor and feature fusion are applied to the sparsely-selected keyframes only. Hence, the computational burden brought by the information-refill mechanism is insignificant.

While information-refill inherits the idea of keyframes, we remark here that unlike existing interval-based methods~\cite{jain2019accel,zhang2017fast} that isolate the intervals for independent processing, the intervals (separated by the keyframes) in \mbox{IconVSR} are connected to maintain a global information propagation.

\begin{table*}[!t]
	\caption{\textbf{Quantitative comparison (PSNR/SSIM).} All results are calculated on Y-channel except REDS4~\cite{nah2019ntire} (RGB-channel). \rf{Red} and \bd{blue} colors indicate the best and the second-best performance, respectively. Blanked entries correspond to results unable to be reported. The runtime is computed on an LR size of $180{\times}320$.}
	\label{tab:quan}
	\vspace{-0.4cm}
	\begin{center}\scalebox{0.84}{
			\tabcolsep=0.1cm
			\begin{tabular}{l|c|c||c|c|c||c|c|c}
				\hline
				\multirow{2}{*}{}               &            &              & \multicolumn{3}{c||}{BI degradation} & \multicolumn{3}{c}{BD degradation}                                                                                                                                \\ \cline{2-9}
				                                & Params (M) & Runtime (ms) & REDS4~\cite{nah2019ntire}            & Vimeo-90K-T~\cite{xue2019video}    & Vid4~\cite{liu2014bayesian} & UDM10~\cite{yi2019progressive} & Vimeo-90K-T~\cite{xue2019video} & Vid4~\cite{liu2014bayesian} \\ \hline
				Bicubic                         & -          & -            & 26.14/0.7292                         & 31.32/0.8684                       & 23.78/0.6347                & 28.47/0.8253                   & 31.30/0.8687                    & 21.80/0.5246                \\
				VESPCN~\cite{caballero2017real} & -          & -            & -                                    & -                                  & 25.35/0.7557                & -                              & -                               & -                           \\
				SPMC~\cite{tao2017detail}       & -          & -            & -                                    & -                                  & 25.88/0.7752                & -                              & -                               & -                           \\
				TOFlow~\cite{xue2019video}      & -          & -            & 27.98/0.7990                         & 33.08/0.9054                       & 25.89/0.7651                & 36.26/0.9438                   & 34.62/0.9212                    & -                           \\
				FRVSR~\cite{sajjadi2018frame}   & 5.1        & 137          & -                                    & -                                  & -                           & 37.09/0.9522                   & 35.64/0.9319                    & 26.69/0.8103                \\
				DUF~\cite{jo2018deep}           & 5.8        & 974          & 28.63/0.8251                         & -                                  & -                           & 38.48/0.9605                   & 36.87/0.9447                    & 27.38/0.8329                \\
				RBPN~\cite{haris2019recurrent}  & 12.2       & 1507         & 30.09/0.8590                         & 37.07/0.9435                       & 27.12/0.8180                & 38.66/0.9596                   & 37.20/0.9458                    & -                           \\
				EDVR-M~\cite{wang2019edvr}      & 3.3        & 118          & 30.53/0.8699                         & 37.09/0.9446                       & 27.10/0.8186                & 39.40/0.9663                   & 37.33/0.9484                    & 27.45/0.8406                \\
				EDVR~\cite{wang2019edvr}        & 20.6       & 378          & 31.09/0.8800                         & \rf{37.61}/\rf{0.9489}             & \bd{27.35}/\bd{0.8264}      & 39.89/0.9686                   & \bd{37.81}/\bd{0.9523}          & 27.85/0.8503                \\
				PFNL~\cite{yi2019progressive}   & 3.0        & 295          & 29.63/0.8502                         & 36.14/0.9363                       & 26.73/0.8029                & 38.74/0.9627                   & -                               & 27.16/0.8355                \\
				MuCAN~\cite{li2020mucan}        & -          & -            & 30.88/0.8750                         & 37.32/0.9465                       & -                           & -                              & -                               & -                           \\
				TGA~\cite{isobe2020video}       & 5.8        & -            & -                                    & -                                  & -                           & -                              & 37.59/0.9516                    & 27.63/0.8423                \\
				RLSP~\cite{fuoli2019efficient}  & 4.2        & 49           & -                                    & -                                  & -                           & 38.48/0.9606                   & 36.49/0.9403                    & 27.48/0.8388                \\
				RSDN~\cite{isobe2020video1}     & 6.2        & 94           & -                                    & -                                  & -                           & 39.35/0.9653                   & 37.23/0.9471                    & 27.92/0.8505                \\
				RRN~\cite{isobe2020revisiting}  & 3.4        & 45           & -                                    & -                                  & -                           & 38.96/0.9644                   & -                               & 27.69/0.8488                \\ \hline
				\textbf{\mbox{BasicVSR} (ours)} & 6.3        & 63           & \bd{31.42}/\bd{0.8909}               & 37.18/0.9450                       & 27.24/0.8251                & \bd{39.96}/\rf{0.9694}         & 37.53/0.9498                    & \bd{27.96}/\bd{0.8553}      \\
				\textbf{\mbox{IconVSR} (ours)}  & 8.7        & 70           & \rf{31.67}/\rf{0.8948}               & \bd{37.47}/\bd{0.9476}             & \rf{27.39}/\rf{0.8279}      & \rf{40.03}/\rf{0.9694}         & \rf{37.84}/\rf{0.9524}          & \rf{28.04}/\rf{0.8570}      \\ \hline
			\end{tabular}}
		\vspace{-0.5cm}
	\end{center}
\end{table*}

\vspace{0.15cm}
\noindent\textbf{Coupled Propagation.}
In bidirectional settings, features are typically propagated in two opposite directions independently. In this design, the features in each propagation branch are computed based on partial information, from either previous frames or future frames.
To exploit the information in the sequences, we propose a coupled propagation scheme, where the propagation modules are inter-connected. As depicted in Fig.~\ref{fig:iconvsr}(b), in coupled propagation, the features propagated backward $h_i^b$ are taken as inputs in the forward propagation module (\cf~Eq.~\ref{eq:pipeline_basicvsr},~\ref{eq:upsample}):
\begin{equation}
	\begin{split}
		&h_i^b = F_b(x_i, x_{i+1}, h_{i+1}^b),\\
		&h_i^f = F_f(x_i, x_{i-1}, {h_i^b}, h_{i-1}^f),\\
		&y_i = U(h_i^f).
	\end{split}
\end{equation}
With coupled propagation, the forward propagation branch receives information from both past and future frames, leading to features of higher quality and hence better outputs. More importantly, since coupled propagation requires only changes of the branch connections, the performance gain can be obtained without introducing computational overhead.

\begin{figure*}[!t]
	\begin{center}
		\includegraphics[width=0.99\textwidth]{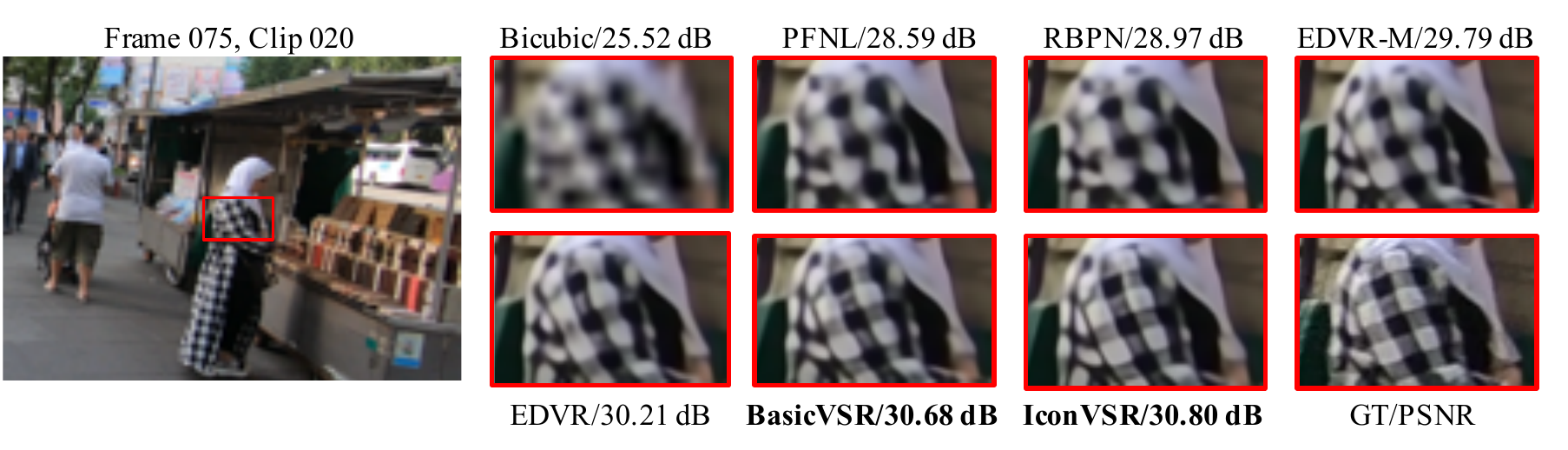}
		\caption{\textbf{Qualitative comparison on REDS4~\cite{nah2019ntire}.} \mbox{BasicVSR} and \mbox{IconVSR} restores clearer square patterns. \mbox{IconVSR} restores sharper edges. \textbf{(Zoom-in for best view)}}
		\vspace{-0.5cm}
		\label{fig:reds4}
	\end{center}
\end{figure*}

\begin{figure*}[!t]
	\begin{center}
		\includegraphics[width=0.99\textwidth]{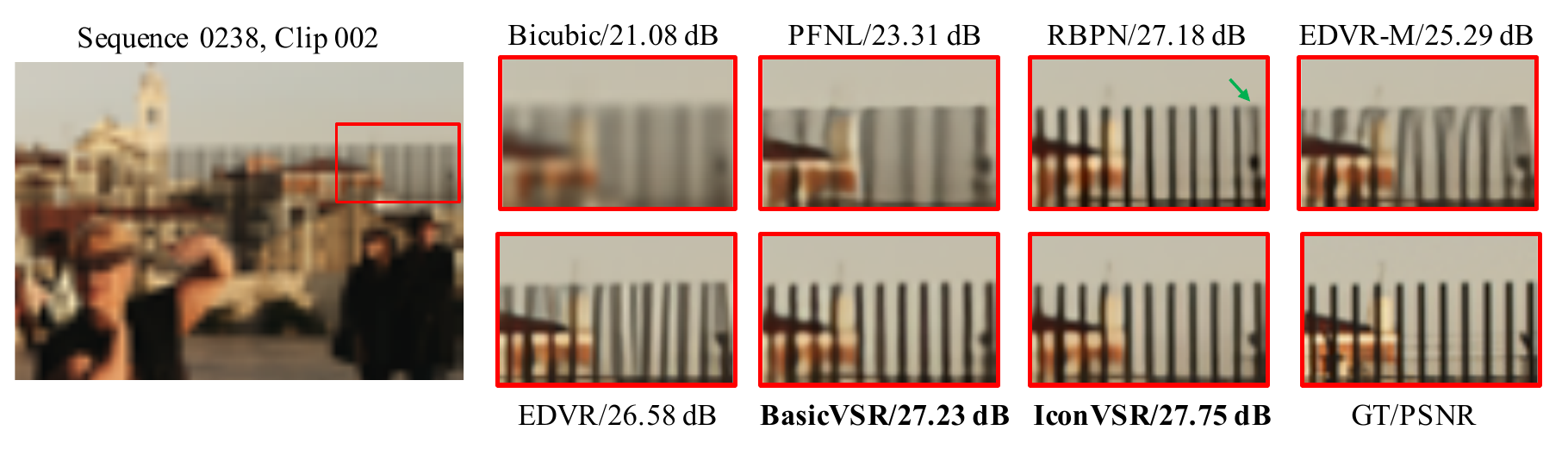}
		\caption{\textbf{Qualitative comparison on Vimeo-90K-T~\cite{xue2019video}.} Only \mbox{BasicVSR} and \mbox{IconVSR} are able to recover the vertical strip patterns. \mbox{IconVSR} restores sharper edges. \textbf{(Zoom-in for best view)}}
		\vspace{-0.5cm}
		\label{fig:vimeo}
	\end{center}
\end{figure*}

\section{Experiments}
\label{sec:exp}
\noindent\textbf{Datasets and Settings}
We consider two widely-used datasets for training: REDS~\cite{nah2019ntire} and Vimeo-90K~\cite{xue2019video}.
For REDS, following~\cite{wang2019edvr}, we use the REDS4 dataset\footnote{Clips 000, 011, 015, 020 of REDS training set.} as our test set. We additionally define REDSval4\footnote{Clips 000, 001, 006, 017 of REDS validation set.} as our validation set. The remaining clips are used for training.
We use Vid4~\cite{liu2014bayesian}, UDM10~\cite{yi2019progressive}, and Vimeo-90K-T~\cite{xue2019video} as test sets along with Vimeo-90K.
We test our models with $4{\times}$ downsampling using two degradations -- Bicubic (BI) and Blur Downsampling (BD).

We use pre-trained SPyNet~\cite{ranjan2017optical} and EDVR-M\footnote{A lightweight version of EDVR.}~\cite{wang2019edvr} as our flow estimation module and feature extractor, respectively. We adopt Adam optimizer~\cite{kingma2014adam} and Cosine Annealing scheme~\cite{loshchilov2016sgdr}. The initial learning rates of the feature extractor and flow estimator are set to $1{\times}10^{-4}$ and $2.5{\times}10^{-5}$, respectively. The learning rate for all other modules is set to $2{\times}10^{-4}$. The total number of iterations is 300K, and the weights of the feature extractor and flow estimator are fixed during the first 5,000 iterations. The batch size is 8 and the patch size of input LR frames is $64{\times}64$. We use Charbonnier loss~\cite{charbonnier1994two} since it better handles outliers and improves the performance over the conventional $\ell_2$ loss~\cite{lai2017deep}. Detailed experimental settings are provided in the appendix.

\subsection{Comparisons with State-of-the-Art Methods}
We conduct comprehensive experiments by comparing \mbox{BasicVSR} and \mbox{IconVSR} with 14 models: VESPCN~\cite{caballero2017real}, SPMC~\cite{tao2017detail}, TOFlow~\cite{xue2019video}, FRVSR~\cite{sajjadi2018frame}, DUF~\cite{jo2018deep}, RBPN~\cite{haris2019recurrent}, EDVR-M~\cite{wang2019edvr}, EDVR~\cite{wang2019edvr}, MuCAN~\cite{li2020mucan}, PFNL~\cite{yi2019progressive}, RLSP~\cite{fuoli2019efficient}, TGA~\cite{isobe2020video}, RSDN~\cite{isobe2020video1}, and RRN~\cite{isobe2020revisiting}. The quantitative results are summarized in Table~\ref{tab:quan} and the speed and performance comparison is provided in Fig.~\ref{fig:teaser}. Note that the parameters of \mbox{BasicVSR} and IconVSR are inclusive of that in the optical flow network, SPyNet. So the comparison is fair.

\vspace{0.15cm}
\noindent\textbf{\mbox{BasicVSR}.} \mbox{BasicVSR} outperforms existing state of the arts on various datasets, including REDS4, UDM10, and Vid4.
\mbox{BasicVSR} also demonstrates high efficiency in addition to improvements in restoration quality. As shown in Fig.~\ref{fig:teaser}, \mbox{BasicVSR} surpasses RSDN~\cite{isobe2020video1} by 0.61~dB on UDM10 while having a similar number of parameters.
When compared with EDVR~\cite{wang2019edvr}, which has a significantly larger complexity, \mbox{BasicVSR} obtains a marked improvement of 0.33~dB on REDS4 and competitive performances on Vimeo-90K-T and Vid4.
We note that the performance of \mbox{BasicVSR} on Vimeo-90K-T is slightly lower than that achieved by sliding-window methods such as EDVR~\cite{wang2019edvr} and TGA~\cite{isobe2020video}. This is expected since Vimeo-90K-T contains sequences with only seven frames, while the success of \mbox{BasicVSR} partially comes from the aggregation of long-term information (which is a realistic assumption).

\noindent\textbf{\mbox{IconVSR}.} \mbox{IconVSR} further improves the performance by up to 0.31~dB over \mbox{BasicVSR} with slightly longer runtime. The performance gain is especially obvious in Vimeo-90K-T and REDS4, showing that our proposed coupled propagation and information-refill mechanisms are beneficial in videos (1) lacking long-term information (Vimeo-90K-T) and (2) containing large and complicated motions (REDS4).
Overall, both \mbox{BasicVSR} and \mbox{IconVSR} are able to achieve remarkable performance while being faster than most state of the arts.

Qualitative comparisons are shown in Figures~\ref{fig:reds4} and \ref{fig:vimeo}. \mbox{BasicVSR} and \mbox{IconVSR} are able to recover finer details and sharper edges. For instance, only \mbox{BasicVSR} and \mbox{IconVSR} successfully recover clear square patterns in Fig.~\ref{fig:reds4} and the vertical strip patterns in Fig.~\ref{fig:vimeo}. With the proposed components, \mbox{IconVSR} is able to reconstruct images with sharper edges. More examples are provided in the appendix.


\section{Ablation Studies}
\label{sec:ablation}
\begin{figure}[!t]
	\begin{center}
		\hspace{-0.3cm}\includegraphics[width=0.49\textwidth]{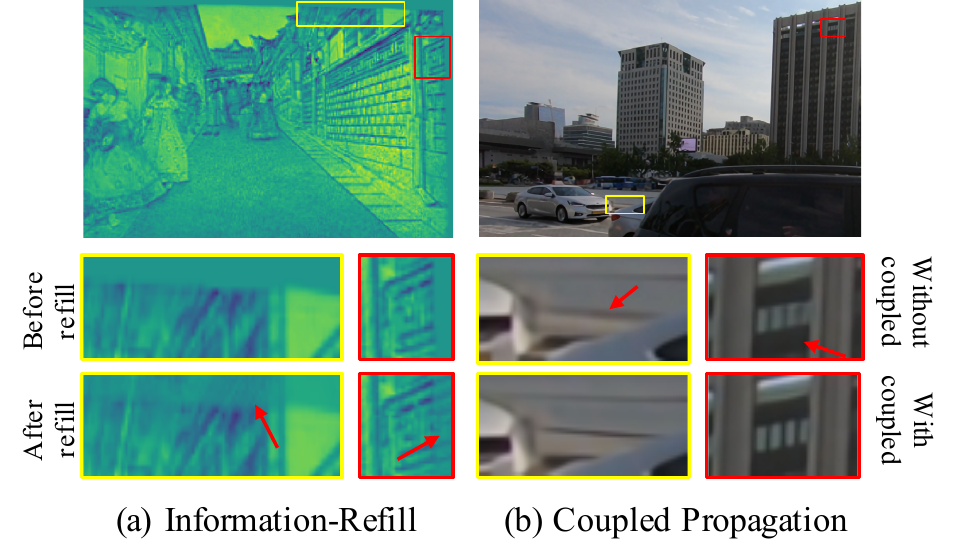}
		\caption{\textbf{(a)} Information lost during spatial warping can be compensated by the additional features. \textbf{(b)} With more effective use of the backward-propagated features, coupled propagation leads to clearer details and finer edges, especially in regions that are occluded in previous frames and regions that exist in the whole sequence. \textbf{(Zoom-in for best view)}}
		\vspace{-0.6cm}
		\label{fig:icon}
	\end{center}
\end{figure}
\begin{figure}[!t]
	\begin{center}
		\includegraphics[width=0.45\textwidth]{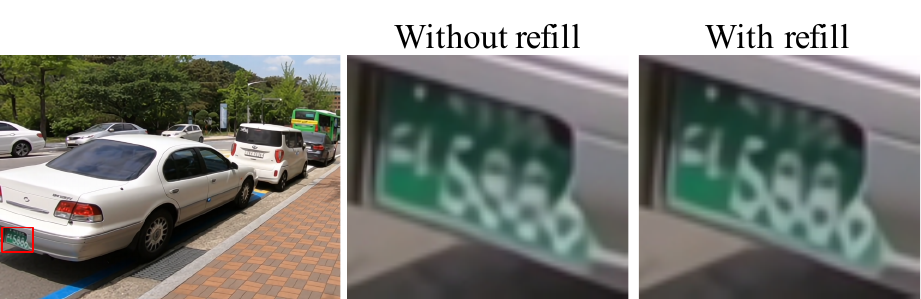}
		\caption{\textbf{Effect of Information-Refill.} The contribution of information-refill is more obvious in regions with fine details, where alignment is error-prone. The information from the additional feature extractor leads to marked improvements.}
		\vspace{-0.6cm}
		\label{fig:refill_qual}
	\end{center}
\end{figure}
\subsection{From \mbox{BasicVSR} to \mbox{IconVSR}}
\label{sec:basic-vs-icon}
\noindent\textbf{Information-Refill.}
We qualitatively visualize the features before and after information-refill to gain insights into the mechanism. As shown in Fig.~\ref{fig:icon}(a), before information-refill, the boundary pixels in the warped feature essentially become zero due to non-existing correspondences. The lost information inevitably worsens the feature quality, leading to degraded outputs.
With our information-refill mechanism, the additional features can be used to ``refill'' the lost information in regions where the features are poorly aligned. The retrieved information can then be employed for the subsequent feature refinement and propagation.

The above effect is especially obvious in regions with fine details. In those regions, information from neighboring frames cannot be effectively aggregated due to alignment error, often resulting in inferior quality. With information-refill, the additional features assist in the restoration of the details, leading to in improved quality.
For example, as shown in Fig.~\ref{fig:refill_qual}, the license plate number can be reconstructed more clearly with the refill mechanism.\\
\vspace{-0.3cm}

\noindent\textbf{Coupled Propagation.}
To ablate the coupled propagation scheme, we disable the information-refill mechanism and compare \mbox{IconVSR} with \mbox{BasicVSR}. In Fig.~\ref{fig:icon}(b), the yellow box represents a region occluded in previous frames, and the forward propagation branch in \mbox{BasicVSR} could not receive information of that region. The red box denotes a region that exists in all frames of the sequence, and hence abundant ``snapshots'' of the region can be found in latter frames.
With coupled propagation, the backward-propagated features are employed more effectively, and hence more details and finer edges can be reconstructed. The PSNR improvement over \mbox{BasicVSR} is summarized in Table~\ref{tab:proto-vs-ir}.

\begin{table}[!t]
	\caption{\textbf{Evaluations of \mbox{IconVSR} components.} The two components bring an improvement of up to 0.28~dB over \mbox{BasicVSR}. The PSNR is computed on REDS4/REDSval4.}
	\label{tab:proto-vs-ir}
	\begin{center}
		\scalebox{0.8}{
			\tabcolsep=0.09cm
			\begin{tabular}{l|c c c}
				             & \mbox{BasicVSR} & \mbox{IconVSR} (w/o refill) & \mbox{IconVSR} \\\hline
				Info-Refill  & \xmark          & \xmark                      & \cmark         \\
				Coupled-Prop & \xmark          & \cmark                      & \cmark         \\\hline
				PSNR         & 31.42/30.17     & 31.60/30.38                 & 31.67/30.45    \\\hline
			\end{tabular}}
	\end{center}
	\vspace{-0.5cm}
\end{table}
\begin{figure}[!t]
	\begin{center}
		\includegraphics[width=0.46\textwidth]{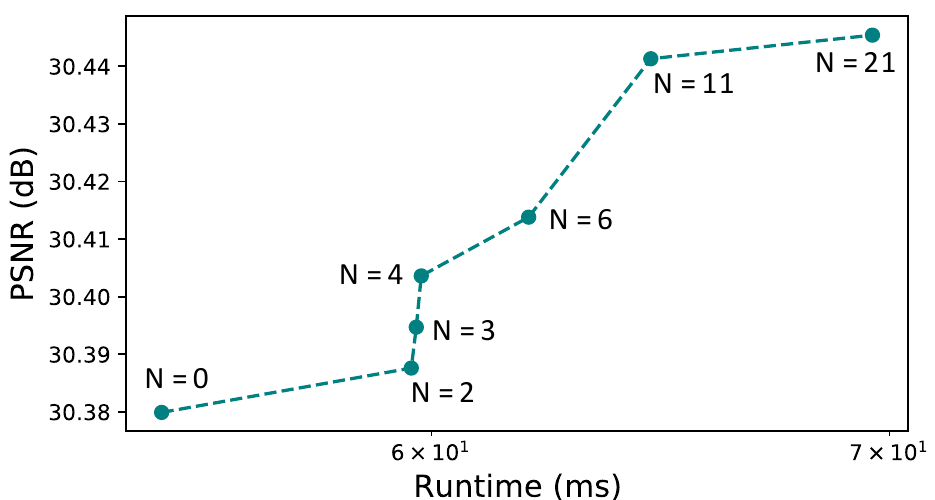}
		\caption{\textbf{Tradeoff in \mbox{IconVSR}.} One can reduce the number of keyframes for faster inference. The PSNR is positively correlated with the number of keyframes $N$, indicating the effectiveness of the information-refill mechanism. The PSNR is calculated on REDSval4. The total number of frames in each clip is 100, and the keyframes are evenly spaced.}
		\vspace{-0.7cm}
		\label{fig:keyframes}
	\end{center}
\end{figure}

\subsection{Tradeoff in \mbox{IconVSR}}
Although \mbox{IconVSR} is trained with a fixed keyframe interval, one can reduce the number of keyframes for faster inference. The PSNR using different numbers of keyframes is depicted in Fig.~\ref{fig:keyframes}, where we see that the PSNR is positively correlated with the number of keyframes, verifying the contributions of the information-refill mechanism.
In an extreme case when there is no keyframe, \mbox{IconVSR} degenerates to a recurrent network. Nevertheless, it still achieves a PSNR of 30.38~dB on REDSval4, which is 0.21~dB higher than \mbox{BasicVSR}. This demonstrates the effectiveness of our coupled propagation scheme, which can be used without introducing additional computational overhead.

\vspace{-0.15cm}
\section{Conclusion}
This work devotes attention to the search of generic and efficient VSR baselines to ease the analysis and extension of VSR approaches. Through decomposing and analyzing existing elements, we propose \mbox{BasicVSR}, a simple yet effective network that outperforms existing state of the arts with high efficiency. We build upon \mbox{BasicVSR} and propose \mbox{IconVSR} with two novel components to further improve the performance. \mbox{BasicVSR} and \mbox{IconVSR} can serve as strong baselines for future works, and the discovery on the architecture designs could potentially be extended to other low-level vision tasks, such as video deblurring, denoising and colorization.

\vspace{0.2cm}
\noindent\textbf{Acknowledgement}.
This research was conducted in collaboration with SenseTime and supported by the Singapore Government through the Industry Alignment Fund - Industry Collaboration Projects Grant. It is also partially supported by Singapore MOE AcRF Tier 1 (2018-T1-002-056) and NTU SUG.

\clearpage
{\small
	\bibliographystyle{ieee_fullname}
	\bibliography{short,bib.bib}
}
\appendix
\section*{Appendix}
\label{appendix}
\section{Architecture and Experimental Settings}
\noindent\textbf{Architecture.}
In all our models, we adopt SPyNet~\cite{ranjan2017optical} as our flow estimator because of its simplicity and efficiency. We use 30 residual blocks in each propagation branch. The feature channel is set to 64. In IconVSR, we adopt EDVR-M\footnote{A lightweight version of EDVR.}~\cite{wang2019edvr} as the additional feature extractor since it maintains a good balance between efficiency and quality. The complexity of the components are summarized in Table~\ref{tab:params}. BasicVSR and IconVSR share the same flow estimator and main network. The main network is a lightweight network, consisting of only 4.9M parameters.
\begin{table}[!h]
	\caption{\textbf{Model complexity of BasicVSR and IconVSR.}}
	\label{tab:params}
	\begin{center}
		\tabcolsep=0.15cm
		\scalebox{1}{
			\begin{tabular}{l|c|c}
				                         & BasicVSR & IconVSR \\\hline
				\text{Flow Estimator}    & 1.4M     & 1.4M    \\
				\text{Main Network}      & 4.9M     & 4.9M    \\
				\text{Feature Extractor} & -        & 2.4M    \\\hline
				\text{Total}             & 6.3M     & 8.7M    \\
			\end{tabular}}
		\vspace{-0.4cm}
	\end{center}
\end{table}
The flow estimator and feature extractor are fine-tuned together with the main network. In all our experiments, every five frames are selected as keyframes. Note that the feature extractor is applied to keyframes only. Therefore, the computational burden brought by it is insignificant.

\vspace{0.15cm}
\noindent\textbf{Datasets.}
We consider two widely-used datasets for training: REDS~\cite{nah2019ntire} and Vimeo-90K~\cite{xue2019video}.
For REDS, following~\cite{wang2019edvr}, we use the REDS4 dataset\footnote{Clips 000, 011, 015, 020 of REDS training set.} as our test set. We additionally define REDSval4\footnote{Clips 000, 001, 006, 017 of REDS validation set.} as our validation set. The remaining clips are used for training.
We use Vid4~\cite{liu2014bayesian}, UDM10~\cite{yi2019progressive}, and Vimeo-90K-T~\cite{xue2019video} as test sets along with Vimeo-90K.

\vspace{0.15cm}
\noindent\textbf{Experimental Settings.}
When training on REDS, we use a sequence of 15 frames as inputs, and loss is computed for the 15 output images. When training on Vimeo-90K, we temporally augment the sequence by flipping the original input sequence to allow longer propagation. In other words, we train with a sequence of 14 frames. During inference, we take the whole video sequence as input.

We adopt Adam optimizer~\cite{kingma2014adam} and Cosine Annealing scheme~\cite{loshchilov2016sgdr}. The initial learning rates of the feature extractor and flow estimator are set to $1{\times}10^{-4}$ and $2.5{\times}10^{-5}$, respectively. The learning rate for all other modules is set to $2{\times}10^{-4}$. The total number of iterations is 300K, and the weights of the feature extractor and flow estimator are fixed during the first 5,000 iterations. The batch size is 8 and the patch size of input LR frames is $64{\times}64$.
\vspace{0.15cm}

\noindent\textbf{Loss Function.}
We use Charbonnier loss~\cite{charbonnier1994two} since it better handles outliers and improves the performance over the conventional $\ell_2$ loss~\cite{lai2017deep}:
\begin{equation}
	\mathcal{L} = \dfrac{1}{N}\sum_{i=0}^N\rho(y_i - z_i),
\end{equation}
where $\rho(x) = \sqrt{x^2 + \epsilon^2}$, $\epsilon{=}1{\times}10^{-8}$, $z_i$ denotes the ground-truth HR frame, and $N$ denotes to the number of pixels.
\vspace{0.15cm}

\noindent\textbf{Degradations.}
We train and test our models with $4{\times}$ downsampling using two degradations -- Bicubic (BI) and Blur Downsampling (BD)~\cite{isobe2020video1,sajjadi2018frame}. For BI, we use the MATLAB function \texttt{imresize} for downsampling. For BD, we blur the ground-truths by a Gaussian filter with $\sigma{=}1.6$, followed by a subsampling every four pixels.
\vspace{0.15cm}

\noindent\textbf{Implementation.}
We implement our models with PyTorch and train the models using two NVIDIA Tesla V100 GPUs. Codes will be made publicly available.


\section{Qualitative Results}
\subsection{Comparison with State of the Arts}
\begin{figure*}[!t]
	\begin{center}
		\includegraphics[width=0.99\textwidth]{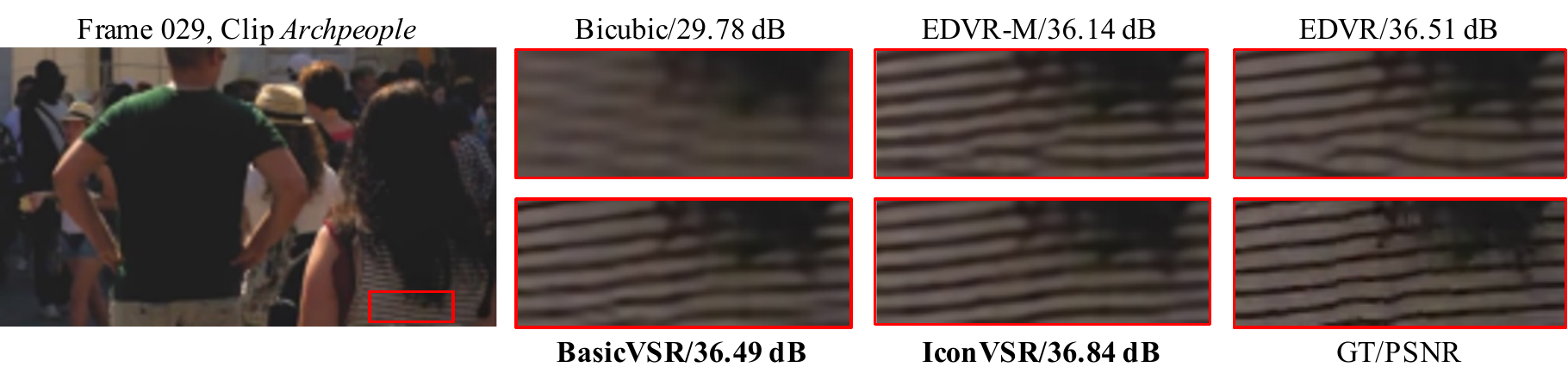}
		\caption{Qualitative comparison on UDM10~\cite{yi2019progressive}.}
		\label{fig:udm10}
	\end{center}
\end{figure*}

In this section, we provide additional qualitative comparisons on REDS4~\cite{nah2019ntire}, Vimeo-90K~\cite{xue2019video}, Vid4~\cite{liu2014bayesian}, and UDM10~\cite{yi2019progressive}. In Fig.~\ref{fig:reds4} to Fig.~\ref{fig:udm10}, it is observed that BasicVSR and IconVSR successfully produce outputs with finer details and sharper edges. Furthermore, with the proposed information-refill and coupled propagation, IconVSR further improves the quality of the outputs.
\begin{figure*}[!t]
	\begin{center}
		\includegraphics[width=0.99\textwidth]{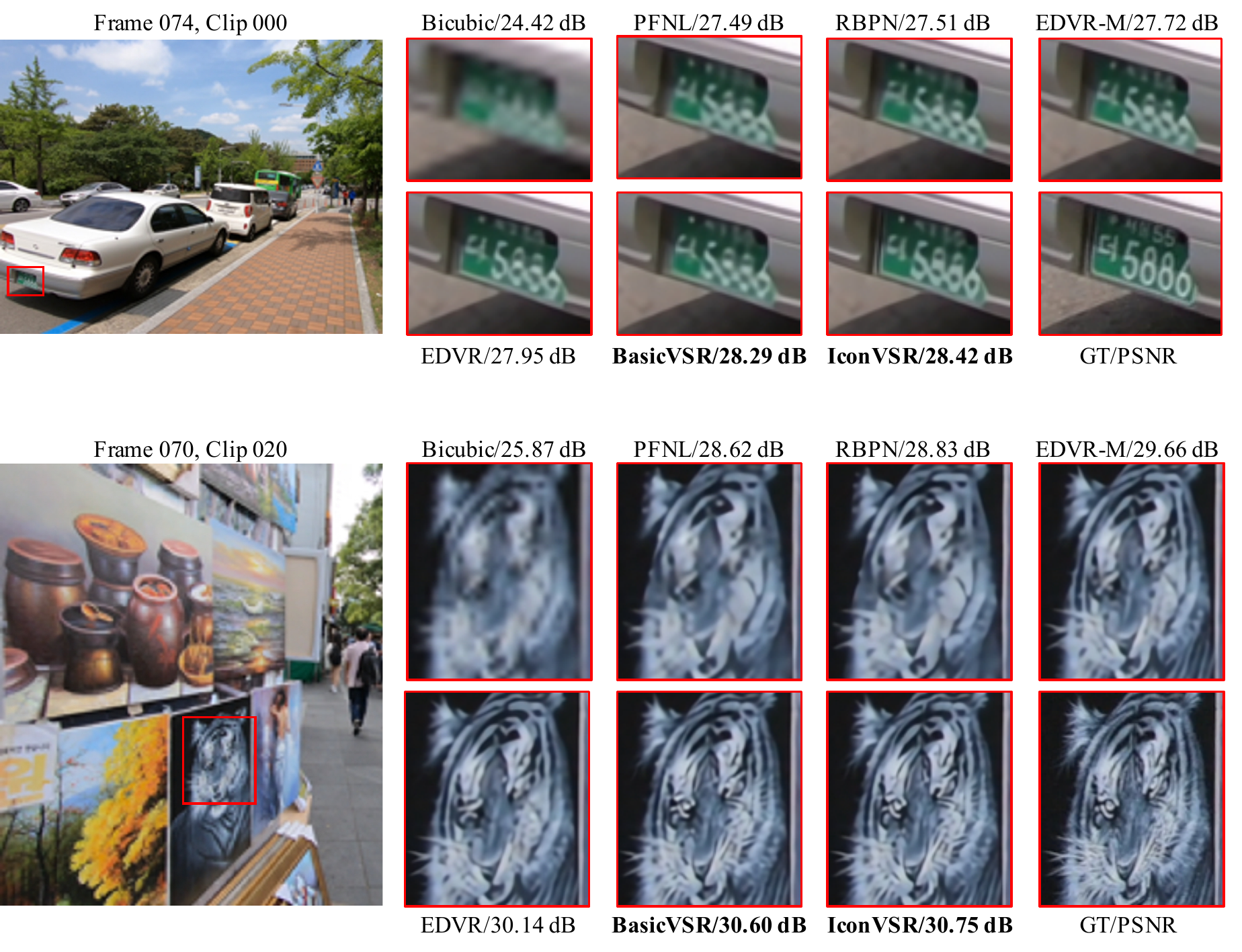}
		\caption{Qualitative comparison on REDS~\cite{nah2019ntire}.}
		\label{fig:reds4}
	\end{center}
\end{figure*}
\begin{figure*}[!t]
	\begin{center}
		\includegraphics[width=0.99\textwidth]{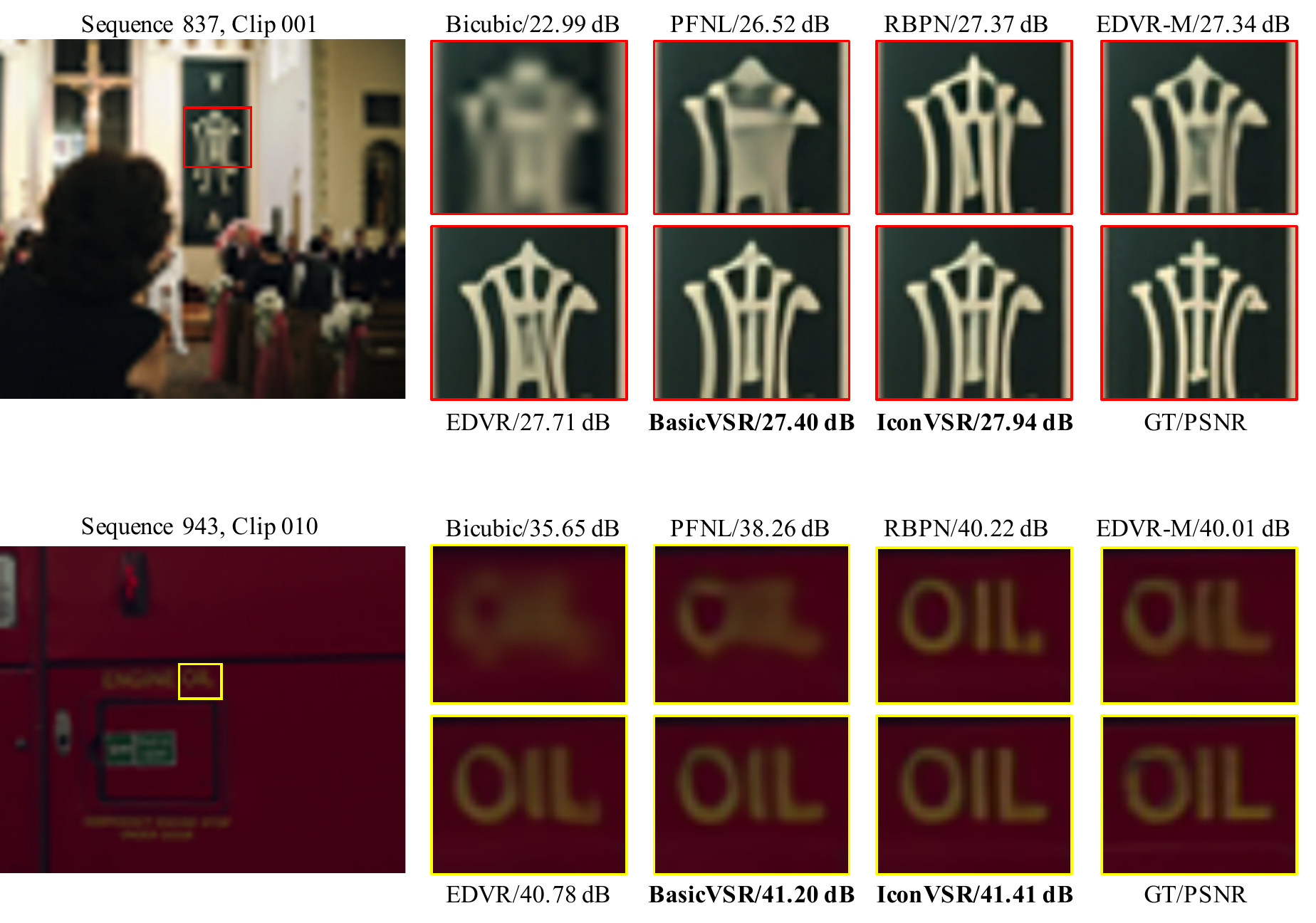}
		\caption{Qualitative comparison on Vimeo-90K~\cite{xue2019video}.}
		\label{fig:vimeo}
	\end{center}
\end{figure*}
\begin{figure*}[!t]
	\begin{center}
		\includegraphics[width=0.99\textwidth]{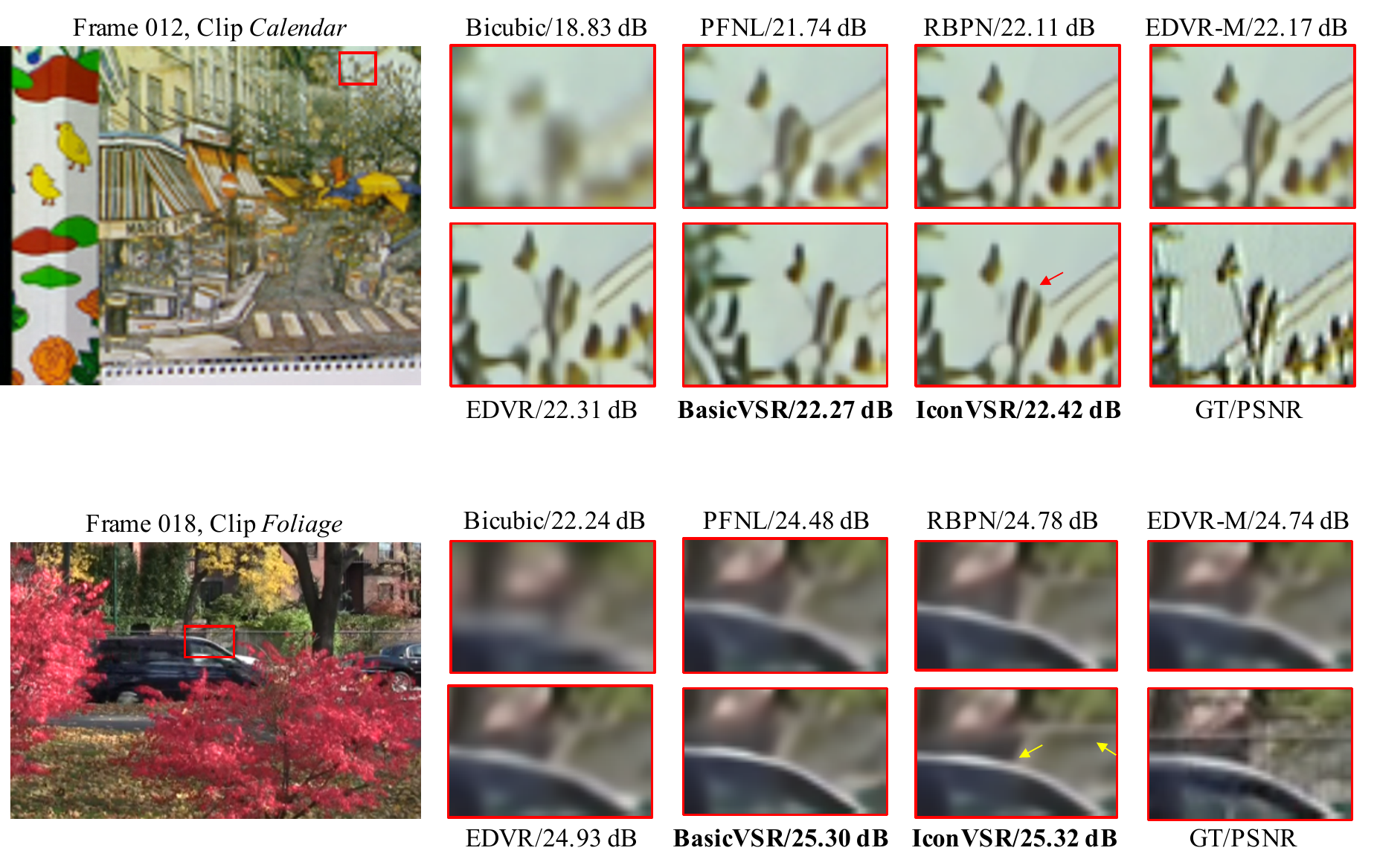}
		\caption{Qualitative comparison on Vid4~\cite{liu2014bayesian}.}
		\label{fig:vid4}
	\end{center}
\end{figure*}
\subsection{BasicVSR vs IconVSR}
\begin{figure*}[!t]
	\begin{center}
		\includegraphics[width=0.99\textwidth]{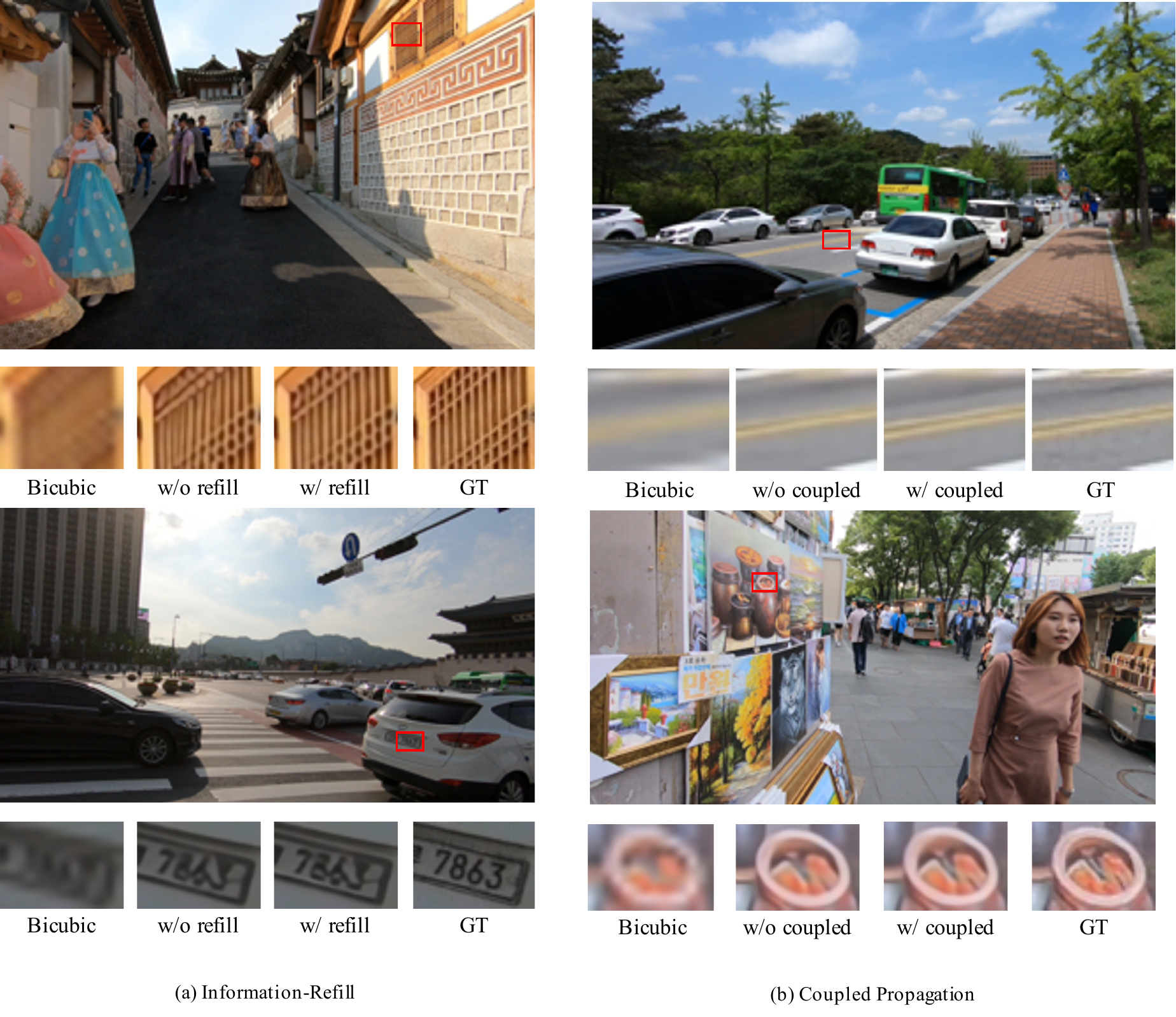}
		\caption{\textbf{Ablation of IconVSR.} With \textit{information-refill} and \textit{coupled propagation}, IconVSR produces outputs with details and sharper edges.}
		\label{fig:ablation}
	\end{center}
\end{figure*}

In Fig.~\ref{fig:ablation}, we provide additional visual comparison of BasicVSR and IconVSR to demonstrate the effectiveness of our proposed components. We see that (1) information-refill improves the output quality on the fine regions, where alignment is error-prone, and (2) coupled propagation leads to sharper edges by better employing the long-term information in the sequence.

\end{document}